\documentclass[sigconf]{acmart}

\usepackage{tabularx}
\usepackage{subcaption}
\usepackage{graphicx}

\AtBeginDocument{%
  \providecommand\BibTeX{{%
    \normalfont B\kern-0.5em{\scshape i\kern-0.25em b}\kern-0.8em\TeX}}}

\setcopyright{acmcopyright}
\copyrightyear{2022}
\acmYear{2022}
\acmDOI{XXXXXXX.XXXXXXX}

\acmConference[GLB '22]{Graph Learning Benchmark}{April 25--29,
  2022}{Lyon, France}
\acmPrice{15.00}
\acmISBN{978-1-4503-XXXX-X/18/06}

\begin{document}

\title{PDNS-Net: A Large Heterogeneous Graph Benchmark Dataset of Network Resolutions for Graph Learning}

\author{Udesh Kumarasinghe}
\email{udeshk@scorelab.org}
\affiliation{%
  \institution{University of Colombo}
  \country{Sri Lanka}}

\author{Fatih Deniz, Mohamed Nabeel}
\email{(fdeniz, mnabeel)@hbku.edu.qa}
\affiliation{%
  \institution{Qatar Computing Research Institute}
  \country{Qatar}
}

\renewcommand{\shortauthors}{Kumarage, et al.}

\begin{abstract}

In order to advance the state of the art in graph learning algorithms, it is necessary to construct large real-world datasets. While there are many benchmark datasets for homogeneous graphs, only a few of them are available for heterogeneous graphs. Furthermore, the latter graphs are small in size rendering them insufficient to understand how graph learning algorithms perform in terms of classification metrics and computational resource utilization. We introduce, PDNS-Net, the largest public heterogeneous graph dataset containing 447K nodes and 897K edges for the malicious domain classification task. Compared to the popular heterogeneous datasets IMDB and DBLP, PDNS-Net is 38 and 17 times bigger respectively. We provide a detailed analysis of PDNS-Net including the data collection methodology, heterogeneous graph construction, descriptive statistics and preliminary graph classification performance. The dataset is publicly available at https://github.com/qcri/PDNS-Net. Our preliminary evaluation of both popular homogeneous and heterogeneous graph neural networks on PDNS-Net reveals that further research is required to improve the performance of these models on large heterogeneous graphs.
\end{abstract}




\maketitle

\section{Introduction}

The availability of various graph datasets across multiple domains have fueled the accelerated development of graph learning techniques that take into consideration both the graph structure and node/edge attributes to either learn low dimensional representations of graph nodes or classify nodes. However, the research community has identified several limitations with the current benchmark datasets  making it difficult to characterize and differentiate modern graph learning techniques: (1) The size of the graphs in terms of the number of nodes and edges is quite limited and (2) Most of these graphs are homogeneous containing only one type of nodes and one type of edges. 

To address these issues, we introduce a new graph dataset called PDNS-Net, a large-scale heterogeneous graph containing the IP resolutions of Internet domains observed in October 2021 through passive DNS data collection~\cite{Weimer:2005:PDNS}. The graph is constructed from a seed set of malicious domains collected from VirusTotal~\cite{virustotal} and the hosting infrastructure behind these seed domains are extracted from a popular passive DNS repository that passively records most of the domain resolutions occur all around the world~\cite{farsight}. Due to various practical reasons, attackers utilize similar hosting infrastructures to host their domains. The network security research community has utilized the structural properties along with the domain/IP attributes to distinguish malicious domains from benign ones~\cite{hindom, graphbp}. However, the datasets utilized and the experiments carried out in such research works suffer from several limitations: (1) the collection and labeling methodology are not clear, (2) primarily homogeneous GNN models are utilized, (3) the datasets are small and (4) the datasets are not publicly available.

In order to assist researchers and practitioners design and explore various graph learning methods quickly as well as execute legacy GNN models that do not scale with the dataset size, we have created a smaller version of PDNS-Net called mPDNS-Net (miniPDNS-Net) by sampling PDNS-Net~\cite{graph_sampling}. 

Our preliminary results of executing various graph neural network models (both homogeneous and heterogeneous) reveal that (1) larger graphs perform better in the classification task compared to the smaller ones in the heterogeneous setting but surprisingly not in the homogeneous setting, and (2) The classification metrics among homogeneous and heterogeneous graph models are marginal, especially for smaller graphs. Thus, PDNS-Net provides the opportunity for the research community to advance the state of the art on graph learning on large heterogeneous graphs by addressing these observations. Furthermore, PDNS-Net provides new research opportunities to advance the graph learning on heterogeneous graphs in several directions, including imbalanced classification, adversarial robustness and explainability. 

\section{Background}
\subsection{Homogeneous and Heterogeneous Graphs}

\begin{definition}
A \textbf{Graph} $\mathcal{G} = (\mathcal{V}, \mathcal{E})$ denotes data structure, where $\mathcal{V}$ is the vertexes (or node set), and $\mathcal{E} \in (\mathcal{V} \times \mathcal{V})$ denotes the edge set. Let $A \in [0, 1]^{|\mathcal{V}| \times \mathcal{|V|}}$ represent a binary adjacency matrix where $A_{ij} = 1 \; if \; (i,j) \in \mathcal{E}$. 

\end{definition}

A Heterogeneous Graph is a graph structure that represents different entities and their different relationships. A heterogeneous graph can be formally defined as follows.

\begin{definition}
A \textbf{Heterogeneous Graph} \cite{hin_survey:tkde:2017} is defined as a graph $\mathcal{G} = \left (\mathcal{V}, \mathcal{E}, \mathcal{A}, \mathcal{R} \right )$, where $\mathcal{V}$ and $\mathcal{E} \in (\mathcal{V} \times \mathcal{V})$ denote the set of nodes and edges, respectively.  Furthermore, $\mathcal{G}$ is associated with a node type mapping $\phi: \mathcal{V} \to \mathcal{A}$ and an edge type mapping $\psi: \mathcal{E} \to \mathcal{R}$ where the number of node types $|\mathcal{A}| > 1$ or the number of edge types $|\mathcal{R}| > 1$. Edge set $\mathcal{E}$ is also represented as an adjacency matrix $A \in [0, 1]^{|\mathcal{V}| \times |\mathcal{V}| \times |\mathcal{R}|}$ such that $A_r \in [0, 1]^{|\mathcal{V}| \times |\mathcal{V}|}$ is a sub matrix representing the edge of type $r \in \mathcal{R}$.
\end{definition}

\begin{figure}[ht]
    \centering
    \includegraphics[width=0.45\textwidth]{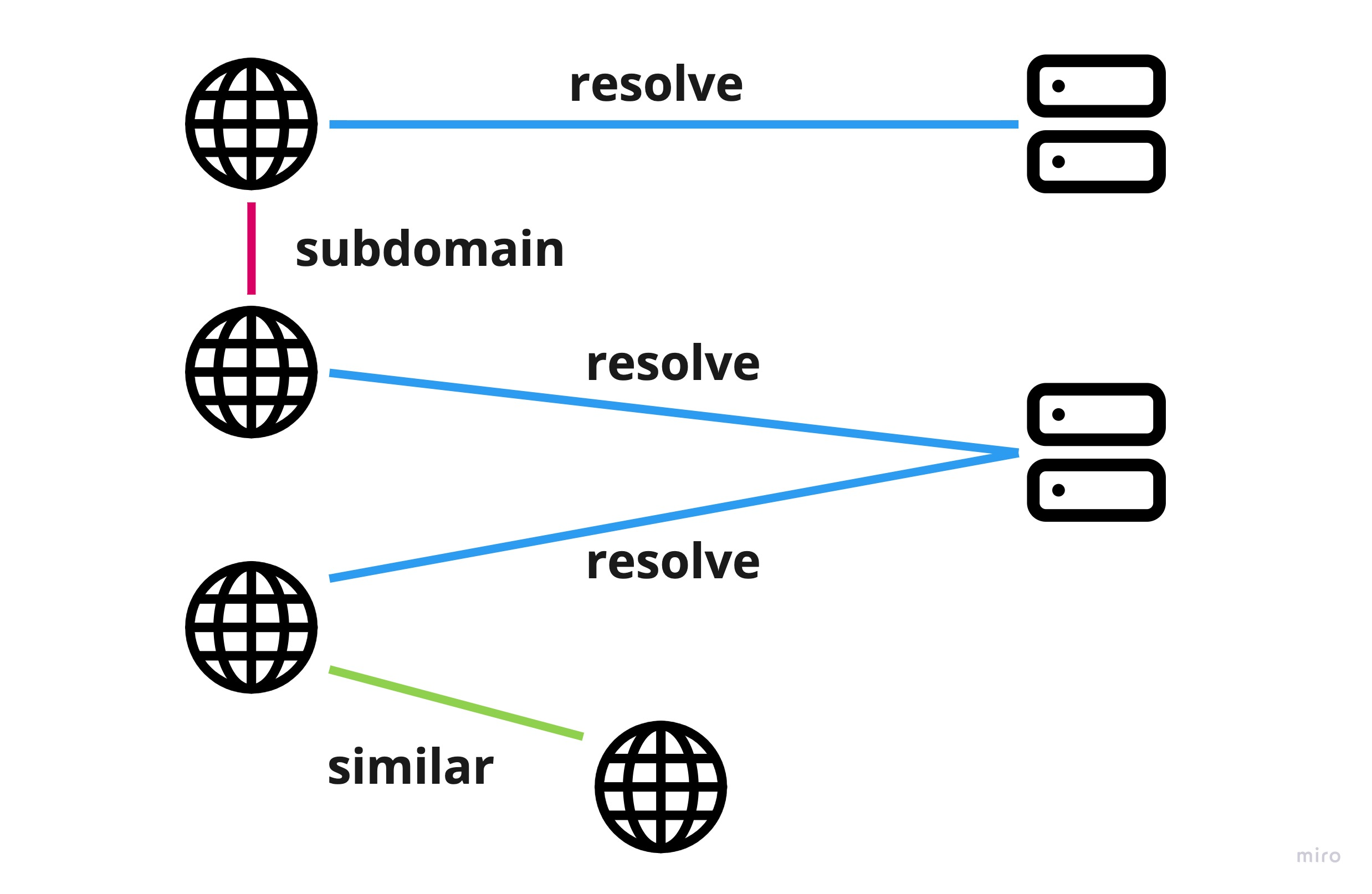}
    \caption{High level schema of the Heterogeneous DNS graph}
    \label{fig:domain_ip_graph}
\end{figure}

In the context of DNS, entities such as domains, or IPs and the  relationships among them are represented as a heterogeneous graph. These relationships between entities can be different connections such as a domain resolving to an IP, or a domain is a subdomain of another. This structure is heterogeneous since it contains either different types of nodes or different types of edges. An advantage of heterogeneity is the ability to represent rich and diverse relationships among the entities. Figure~\ref{fig:domain_ip_graph} shows the high level schema of the PDNS-Net DNS graph.

\subsection{DNS and Malicious Domains}
DNS is a hierarchical naming system that helps map domain names to IP addresses in the Internet. It is one of the core protocol suites of the Internet. 
It provides a distributed database that maps domain name to record sets, such 
as IP addresses. 
We make a distinction between private and public domains. A domain is public if its subdomains or path suffixes
are not created and not under the control of the  domain
owner, whereas a domain is private if its subdomains
are created and managed by the domain owner.

Internet domains are the launchpad for many cyber attacks we
observe nowadays. We term those domains that are utilized in such attacks as malicious domains~\cite{compromised}.

\subsection{Passive DNS}
Passive DNS (PDNS)~\cite{Weimer:2005:PDNS} captures traffic by sensors cooperatively deployed in various DNS hierarchy locations. For example, Farsight PDNS data~\cite{farsight} utilizes sensors deployed behind DNS resolvers and provides aggregate information about domain resolutions. We use Farsight PDNS DB to extract domain resolutions retrospectively.

\section{Data Collection and Characteristics}
\subsection{Data Collection}
\begin{table*}[!ht]
\caption{Lexical Features} 
\label{tab:features}
\begin{minipage}{\linewidth}
\centering
\footnotesize
\begin{tabular}{| p{3cm} | p{10.5cm} | p{3cm} |}
\hline
\multicolumn{1}{|c|}{\textbf{Feature Name}} &  \multicolumn{1}{c|}{\textbf{Description}} &
\multicolumn{1}{c|}{\textbf{Source}} \\ \hline
\#Subdomains
&
The number of levels in the subdomain part of the FQDN
&
\cite{Mabeyondblacklists}
\\ \hline

Minus
&
The number of dashes appear in the FQDN
&
\cite{Mabeyondblacklists}
\\ \hline

Brand
&
Does it impersonate a popular Alexa top 1000 brand?
&
Derived from \cite{Kintis:2017:Combosquatting}
\\ \hline

Similar
&
Does the domain contain words within Levenshtein distance 2 of a popular Alexa top brand?
&
Derived from \cite{Kintis:2017:Combosquatting}
\\ \hline

Fake\_TLD
&
Does the domain name include a fake gTLD (com, edu, net, org, gov)?
&
Derived from \cite{embeddingsquatting:2019}
\\ \hline

Pop\_Keywords
&
Does the domain name include popular keywords?
&
Derived from \cite{Kintis:2017:Combosquatting}
\\ \hline

Entropy
&
The entropy of the FQDN
&
\cite{dgamal:usenix:2012,Bahnsen:2017:PhishingURLsNN}
\\ \hline
\end{tabular}
\end{minipage}
\vspace{-4mm}
\end{table*}

We obtain ground truth data and passive DNS resolutions to build the DNS graph from the data collected from different sources during the period of 11/10/2020 to 18/10/2020.

\begin{figure*}[ht]
\includegraphics[width=0.85\textwidth]{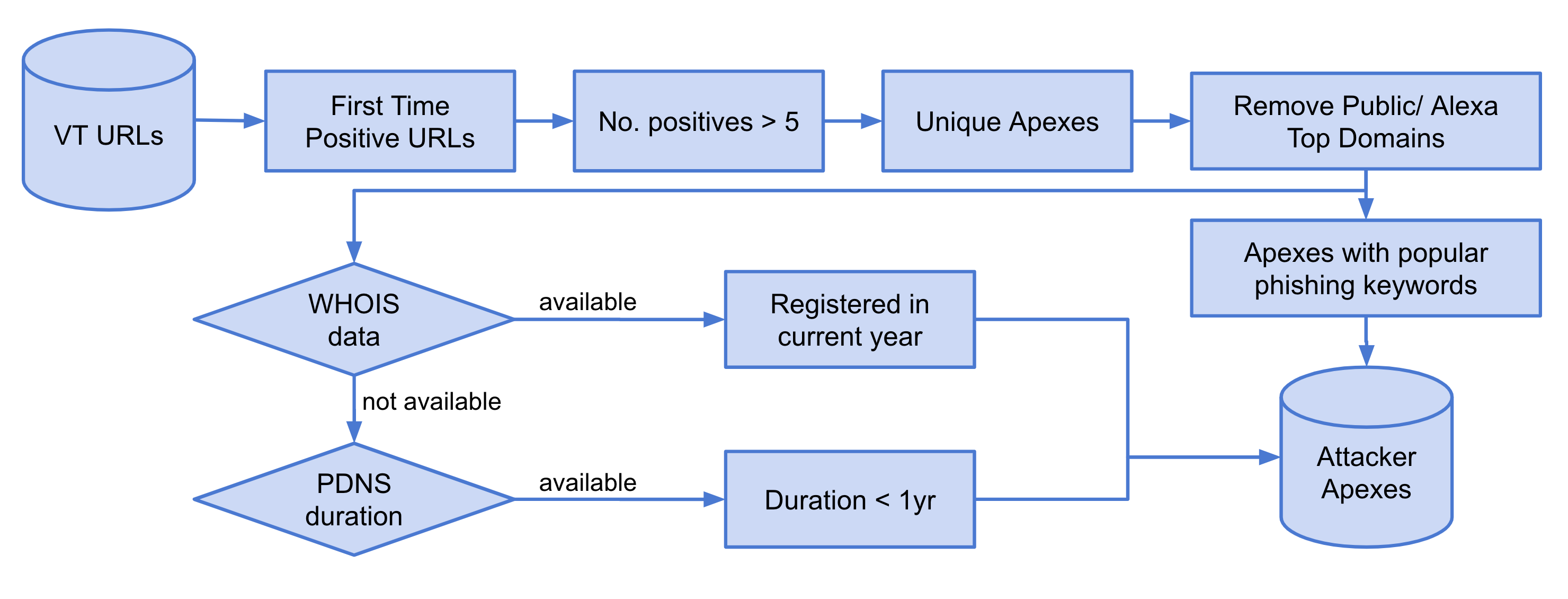}
\caption{Groud truth malicious domain filtering pipeline}
\centering
\label{fig:filtering}
\end{figure*}

\textbf{Malicious Ground Truth}. As shown in Figure~\ref{fig:filtering}, we collect malicious domain ground truth from VirusTotal Feed~\cite{virustotal}. We execute this pipeline to identify malicious domains created by attackers as they exhibit homophily relationships.

\textbf{Benign Ground Truth}. Our benign ground truth is collected from the Alexa top 1m list~\cite{alexa}. 
Although some malicious domains make to the top domain list if they attract high popularity, such domains do not persist on the top domain list over a period of time. Thus, following prior research outcomes~\cite{graphbp}, we select those domains in Alexa top 1m that consistently appear for 90 days as our benign ground truth. 

\subsection{Passive DNS Expansion}
We expand the malicious seed identified from the VirusTotal feed to identify domain-IP resolutions using the Farsight PDNS service~\cite{farsight}.
The following steps are performed to collect the passive DNS resolutions: 

\begin{enumerate}
    \item Collect the IPs hosting the malicious domains.
    \item Collect the domains hosted on IPs collected in step 1.
    \item Collect the IPs hosting the domains collected in step 2.
\end{enumerate}

\subsection{Graph Construction}

The collected domain-IP resolutions are then used to build a heterogeneous knowledge graph consisting of nodes of type domain, IP, subdomain and 3 types of relationships among them.

\textbf{domain-resolve-ip}. The relationship represents the domains hosted on a given IP.

\textbf{domain-similar-domain}. The character level similarity between two domains is used to generate relationships between domain nodes. In order to find domain similarity, n-gram is used to process domain names and embed each domain in a high level representation using TF-IDF vectorization. Then, the cosine similarity is used to identify domains with character level similarity. Tri-grams are used for the n-gram process and 0.8 similarity threshold is used for cosine similarity.

\textbf{domain-subdomain-domain}. The relationship represents the domains sharing the same apex domain.

\subsection{Node Feature Extraction and Metapaths}

The lexical features of domain nodes in Table~\ref{tab:features} capture the lexical formation of domain names. Attackers create many domains to remain agile but such domains leave certain revealing patterns allowing one to associate them by their lexical features. These features are extracted by pre-processing the domain names before building the graph. Subnet and ASN of each IP are used as categorical features of the IP nodes. An added advantage of these features is that they are quite efficient to collect. 

We define the following metapaths~\cite{metapath2vec} for the PDNS-Net graph:
\begin{itemize}
    \item domain - similar - domain
    \item domain - subdomain - domain
    \item domain - resolve - IP - resolve - domain
\end{itemize}

\subsection{Graph Pruning}
We perform the following pruning based on empirically identified thresholds in order to reduce the noise in the graph.

\textbf{IP Pruning}. 
IP nodes having higher degrees are most likely from firewalls or some public IPs. Thus, domains hosted on such IPs are less likely to be related to one another. We prune IPs hosting more than 1500 domains.

\textbf{Public domain removal}. Public domains (i.e. wix.com or 000webhostapp.com) host many unrelated subdomains. Two subdomains sharing the same public domain are less likely to be related. Thus, we prune all public domains from the resolution graph. 


\textbf{Isolated node removal}. We prune those connected components having only one domain node as they do not contribute to any graph learning algorithm.

\subsection{mPDNS-Net Dataset}
The majority of current graph adversarial attacks~\cite{attackscale2021} and many of the advanced GNN algorithms~\cite{gat, hgt} are inefficient on large graphs such as ours. Therefore, a representative subgraph of PDNS-Net is generated to be compatible with current graph adversarial attacks and GNN models. Multiple methods have been researched on sampling representative subgraphs from large-scale graphs. According to~\cite{graph_sampling}, exploratory sampling strategies based on random-walks outperforms both uniform node sampling and edge selection based strategies. Thus, an exploratory random-walk graph sampling method, Metropolis-Hastings graph sampling algorithm~\cite{mh_graph_sampling}, is used to generate a representative subgraph of the PDNS-Net graph. We refer to the sample dataset as mPDNS-Net.

\subsection{Descriptive Statistics}
\begin{figure*} [!ht]
  \begin{subfigure}[t]{0.33\textwidth}
    \includegraphics[width=\textwidth]{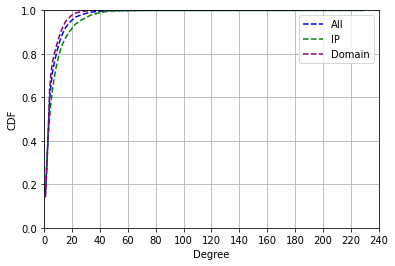}
    \caption{mDNS}
    \label{fig:mdns_degree}
  \end{subfigure}\hfill
  \begin{subfigure}[t]{0.33\textwidth}
    \includegraphics[width=\textwidth]{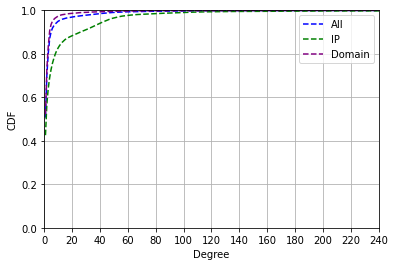}
    \caption{DNS}
    \label{fig:dns_degree}
  \end{subfigure}
  \begin{subfigure}[t]{0.33\textwidth}
    \includegraphics[width=\textwidth]{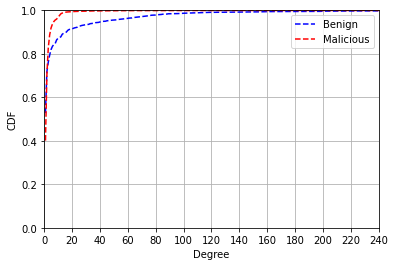}
    \caption{Benign vs. Malicious}
    \label{fig:benmal_degree}
  \end{subfigure}\hfill
  \caption{Degree distribution of the two datasets}
  \label{fig:degree}
\end{figure*}

Table~\ref{tab:graphstats} shows the first order statistics of the two datasets. Figure~\ref{fig:degree} shows the node degree distributions for the two datasets. We make several observations from these statistics: (1) Both DNS and mDNS datasets have similar node degree distributions confirming that our sampling approach is representative, (2) The degrees of IP nodes are higher than that of domains, and (3) Benign domains are hosted on more IPs than malicious domains.

\begin{table}[!th]
\small
\begin{center}
\caption{Graph statistics of the datasets}~\label{tab:graphstats}
\begin{tabular}{|c|c|c|c|c|c|}\hline
\textbf{Dataset}  & \textbf{\#Domains} & \textbf{\#IPs} & \textbf{\#Edges} & \textbf{\#Malicious} & \textbf{\#Benign}\\
\hline
mDNS &  7,495 & 4,505 & 37,285 & 2,827 & 4,668 \\
\hline
DNS & 373,475 & 73,593 & 897,588 & 20,354 & 4,963 \\
\hline
\end{tabular}

\end{center}
\end{table}

\section{Node Classification Results}


\begin{table*}[!ht]
  \caption{Malicious domain detection performance comparison using different GNN models on the two datasets}
  \label{tab:results_hetero}
  \begin{tabular}{|l||c|c|c|c|c|c|c|c|c|c|c|c||}
    \toprule
    Methods & \multicolumn{6}{c|}{mDNS} & \multicolumn{6}{c|}{DNS} \\ 
    \cline{2-13} 
    & Acc. & AUC & F1 & Prec. & Recall & FPR & Acc. & AUC & F1 & Prec. & Recall & FPR \\
    \midrule
    GCN & 0.81 & 0.81 & 0.81 & 0.81 & 0.80 & 0.18 & 0.77 & 0.77 & 0.77 & 0.79 & 0.72 & 0.18 \\
    GraphSage & 0.84 & 0.84 & 0.84 & 0.80 & 0.90 & 0.22 & 0.76 & 0.76 & 0.76 & 0.80 & 0.71 & 0.18 \\
    GAT & 0.83 & 0.83 & 0.83 & 0.82 & 0.85 & 0.20 & 0.76 & 0.76 & 0.76 & 0.76 & 0.74 & 0.23 \\
    \hline
    RGCN & 0.85 & 0.85 & 0.85 & 0.84 & 0.86 & 0.16 & 0.78 & 0.78 & 0.78 & 0.81 & 0.74 & 0.17 \\
    HGT & 0.87 & 0.87 & 0.87 & 0.88 & 0.86 & 0.12 & 0.90 & 0.90 & 0.90 & 0.92 & 0.88 & 0.08 \\
    HeteroSAGE & 0.89 & 0.89 & 0.89 & 0.91 & 0.86 & 0.09 & 0.93 & 0.93 & 0.93 & 0.94 & 0.92 & 0.06 \\
    HeteroGAT & 0.86 & 0.86 & 0.86 & 0.84 & 0.89 & 0.17 & 0.94 & 0.90 & 0.94 & 0.96 & 0.97 & 0.17 \\
  \bottomrule
\end{tabular}
\end{table*}

Some GNNs are designed for homogeneous graphs that consist of only one type of node and one type of edge, such as GCN~\cite{gcn}, GraphSAGE~\cite{graphsage}, and GAT~\cite{gat}. Some others are designed for heterogeneous graphs and can deal with different types of nodes and relationships, such as  HGT~\cite{hgt}, while others can handle only different types of relationships, but not node types, like RGCN~\cite{rgcn}. It is also possible to use generic wrappers, on GNNs that are designed for homogeneous graphs, to deliver messages from source nodes to target nodes based on the bipartite GNN layer for each edge type, and compute graph convolution on heterogeneous graphs. We use such a wrapper in PyG~\cite{Fey:2019:Pyg} environment and generate heterogeneous counterparts of GraphSAGE~\cite{graphsage} and GAT~\cite{gat} and named them as HeteroSAGE and HeteroGAT, respectively. 

In this section, we introduce experimental results on the two datasets described in this paper, namely mPDNS-Net and PDNS-Net, on above-mentioned different variants of GNNs designed for homogeneous and heterogeneous graphs.
We compare their performance according to several metrics including accuracy, area under the curve, F1-score, precision, and false positive rate as shown in Table~\ref{tab:results_hetero}. For heterogeneous to homogeneous conversion, all features with the same feature dimensionality across different types are merged into a single representation and the missing dimensions are filled with zero values. For training the GNN models, we use an Adam optimizer with the learning rate of $5$ x $10^{-3}$, weight decay of $1$ x $10^{-3}$, and the number of epochs of 200. The designed models consist of two GNN layers with the hidden dimension size of 64 and a linear layer for the classification. Models are implemented using PyTorch Geometric (PyG)~\cite{Fey:2019:Pyg} version 2.0.3 built on top of PyTorch version 1.10.0. 

We make several surprising observations that require further attention from the research community. For mPDNS-Net, while heterogeneous models perform better than homogeneous models, the improvement in performance is marginal. An important research direction is to construct heterogeneous models that perform much better on small heterogeneous graphs compare to the homogeneous counterparts. In comparison to the heterogeneous models on the smaller dataset mPDNS-Net, as expected, similar models perform better on the larger dataset PDNS-Net as it is likely to capture richer associations and interactions among domain nodes. However, it is surprising to observe that such performance gain is not achieved for the homogeneous models. We assess that further study is required to diagnose the poor performance of homogeneous models on the large dataset PDNS-Net.

\section{Conclusion}
Graph learning plays a critical role in advancing the machine learning tasks on interconnected graph data. Currently, while there are many publicly available homogeneous graph datasets, only a handful of heterogeneous graph datasets are available albeit being small in size. We address this by constructing PDNS-Net, the largest publicly available heterogeneous graph dataset for the malicious domain classification. Our preliminary results on both homogeneous and heterogeneous GNN models on PDNS-Net show that the research community needs to do more to improve the performance of GNNs on such  heterogeneous large datasets. 

\clearpage

\begin{thebibliography}{23}


\ifx \showCODEN    \undefined \def \showCODEN     #1{\unskip}     \fi
\ifx \showDOI      \undefined \def \showDOI       #1{#1}\fi
\ifx \showISBNx    \undefined \def \showISBNx     #1{\unskip}     \fi
\ifx \showISBNxiii \undefined \def \showISBNxiii  #1{\unskip}     \fi
\ifx \showISSN     \undefined \def \showISSN      #1{\unskip}     \fi
\ifx \showLCCN     \undefined \def \showLCCN      #1{\unskip}     \fi
\ifx \shownote     \undefined \def \shownote      #1{#1}          \fi
\ifx \showarticletitle \undefined \def \showarticletitle #1{#1}   \fi
\ifx \showURL      \undefined \def \showURL       {\relax}        \fi
\providecommand\bibfield[2]{#2}
\providecommand\bibinfo[2]{#2}
\providecommand\natexlab[1]{#1}
\providecommand\showeprint[2][]{arXiv:#2}

\bibitem[ale(2022)]%
        {alexa}
 \bibinfo{year}{2022}\natexlab{}.
\newblock \bibinfo{title}{{Alexa: The Web Information Company}}.
\newblock \bibinfo{howpublished}{\url{https://www.alexa.com/topsites}}.
\newblock
\newblock
\shownote{Accessed January 2022}.


\bibitem[Antonakakis et~al\mbox{.}(2012)]%
        {dgamal:usenix:2012}
\bibfield{author}{\bibinfo{person}{M. Antonakakis}, \bibinfo{person}{R.
  Perdisci}, \bibinfo{person}{Y. Nadji}, \bibinfo{person}{N. Vasiloglou},
  \bibinfo{person}{S. Abu-Nimeh}, \bibinfo{person}{W. Lee}, {and}
  \bibinfo{person}{D. Dagon}.} \bibinfo{year}{2012}\natexlab{}.
\newblock \showarticletitle{From Throw-Away Traffic to Bots: Detecting the Rise
  of DGA-Based Malware}. In \bibinfo{booktitle}{\emph{Presented as part of the
  21st {USENIX} Security}}. \bibinfo{publisher}{{USENIX}},
  \bibinfo{address}{Bellevue, WA}, \bibinfo{pages}{491--506}.
\newblock
\showISBNx{978-931971-95-9}
\urldef\tempurl%
\url{https://www.usenix.org/conference/usenixsecurity12/technical-sessions/presentation/antonakakis}
\showURL{%
\tempurl}


\bibitem[Bahnsen et~al\mbox{.}(2017)]%
        {Bahnsen:2017:PhishingURLsNN}
\bibfield{author}{\bibinfo{person}{A.~C. Bahnsen}, \bibinfo{person}{E.~C.
  Bohorquez}, \bibinfo{person}{S. Villegas}, \bibinfo{person}{J. Vargas}, {and}
  \bibinfo{person}{F.~A. Gonzalez}.} \bibinfo{year}{2017}\natexlab{}.
\newblock \showarticletitle{Classifying phishing URLs using recurrent neural
  networks}. In \bibinfo{booktitle}{\emph{eCrime}}. \bibinfo{pages}{1--8}.
\newblock


\bibitem[Dong et~al\mbox{.}(2017)]%
        {metapath2vec}
\bibfield{author}{\bibinfo{person}{Yuxiao Dong}, \bibinfo{person}{Nitesh~V
  Chawla}, {and} \bibinfo{person}{Ananthram Swami}.}
  \bibinfo{year}{2017}\natexlab{}.
\newblock \showarticletitle{metapath2vec: Scalable representation learning for
  heterogeneous networks}. In \bibinfo{booktitle}{\emph{Proceedings of the 23rd
  ACM SIGKDD international conference on knowledge discovery and data mining}}.
  \bibinfo{pages}{135--144}.
\newblock


\bibitem[{Farsight Security, Inc.}(2022)]%
        {farsight}
\bibfield{author}{\bibinfo{person}{{Farsight Security, Inc.}}}
  \bibinfo{year}{2022}\natexlab{}.
\newblock \bibinfo{title}{{DNS Database}}.
\newblock \bibinfo{howpublished}{\url{https://www.dnsdb.info/}}.
\newblock


\bibitem[Fey and Lenssen(2019)]%
        {Fey:2019:Pyg}
\bibfield{author}{\bibinfo{person}{Matthias Fey} {and} \bibinfo{person}{Jan~E.
  Lenssen}.} \bibinfo{year}{2019}\natexlab{}.
\newblock \showarticletitle{Fast Graph Representation Learning with {PyTorch
  Geometric}}. In \bibinfo{booktitle}{\emph{ICLR Workshop on Representation
  Learning on Graphs and Manifolds}}.
\newblock


\bibitem[Geisler et~al\mbox{.}(2021)]%
        {attackscale2021}
\bibfield{author}{\bibinfo{person}{S. Geisler}, \bibinfo{person}{D. Zugner},
  \bibinfo{person}{A. Bojchevski}, {and} \bibinfo{person}{S. Gunnemann}.}
  \bibinfo{year}{2021}\natexlab{}.
\newblock \showarticletitle{Attacking Graph Neural Networks at Scale}. In
  \bibinfo{booktitle}{\emph{Deep Learning for Graphs at AAAI}}.
\newblock


\bibitem[Hamilton et~al\mbox{.}(2017)]%
        {graphsage}
\bibfield{author}{\bibinfo{person}{W. Hamilton}, \bibinfo{person}{Z. Ying},
  {and} \bibinfo{person}{J. Leskovec}.} \bibinfo{year}{2017}\natexlab{}.
\newblock \showarticletitle{Inductive Representation Learning on Large Graphs}.
  In \bibinfo{booktitle}{\emph{NIPS}}.
\newblock


\bibitem[Hu et~al\mbox{.}(2020)]%
        {hgt}
\bibfield{author}{\bibinfo{person}{Ziniu Hu}, \bibinfo{person}{Yuxiao Dong},
  \bibinfo{person}{Kuansan Wang}, {and} \bibinfo{person}{Yizhou Sun}.}
  \bibinfo{year}{2020}\natexlab{}.
\newblock \showarticletitle{Heterogeneous graph transformer}. In
  \bibinfo{booktitle}{\emph{Proceedings of The Web Conference 2020}}.
  \bibinfo{pages}{2704--2710}.
\newblock


\bibitem[Hubler et~al\mbox{.}(2008)]%
        {mh_graph_sampling}
\bibfield{author}{\bibinfo{person}{C. Hubler}, \bibinfo{person}{H. Kriegel},
  \bibinfo{person}{K. Borgwardt}, {and} \bibinfo{person}{Z. Ghahramani}.}
  \bibinfo{year}{2008}\natexlab{}.
\newblock \showarticletitle{Metropolis Algorithms for Representative Subgraph
  Sampling}. In \bibinfo{booktitle}{\emph{ICDM}}. \bibinfo{pages}{283--292}.
\newblock


\bibitem[Kintis et~al\mbox{.}(2017)]%
        {Kintis:2017:Combosquatting}
\bibfield{author}{\bibinfo{person}{P. Kintis}, \bibinfo{person}{N.
  Miramirkhani}, \bibinfo{person}{C. Lever}, \bibinfo{person}{Y. Chen},
  \bibinfo{person}{R. Romero-Gomez}, \bibinfo{person}{N. Pitropakis},
  \bibinfo{person}{N. Nikiforakis}, {and} \bibinfo{person}{M. Antonakakis}.}
  \bibinfo{year}{2017}\natexlab{}.
\newblock \showarticletitle{Hiding in Plain Sight: A Longitudinal Study of
  Combosquatting Abuse}. In \bibinfo{booktitle}{\emph{CCS}}.
  \bibinfo{publisher}{ACM}, \bibinfo{address}{New York, NY, USA},
  \bibinfo{pages}{569--586}.
\newblock


\bibitem[Kipf and Welling(2017)]%
        {gcn}
\bibfield{author}{\bibinfo{person}{T. Kipf} {and} \bibinfo{person}{M.
  Welling}.} \bibinfo{year}{2017}\natexlab{}.
\newblock \showarticletitle{{Semi-Supervised Classification with Graph
  Convolutional Networks}}. In \bibinfo{booktitle}{\emph{ICLR}}.
\newblock


\bibitem[Leskovec and Faloutsos(2006)]%
        {graph_sampling}
\bibfield{author}{\bibinfo{person}{J. Leskovec} {and} \bibinfo{person}{C.
  Faloutsos}.} \bibinfo{year}{2006}\natexlab{}.
\newblock \showarticletitle{Sampling from Large Graphs}. In
  \bibinfo{booktitle}{\emph{KDD}}. \bibinfo{pages}{631‚Äì636}.
\newblock


\bibitem[Ma et~al\mbox{.}(2009)]%
        {Mabeyondblacklists}
\bibfield{author}{\bibinfo{person}{Justin Ma}, \bibinfo{person}{Lawrence~K.
  Saul}, \bibinfo{person}{Stefan Savage}, {and} \bibinfo{person}{Geoffrey~M.
  Voelker}.} \bibinfo{year}{2009}\natexlab{}.
\newblock \showarticletitle{Beyond Blacklists: Learning to Detect Malicious Web
  Sites from Suspicious URLs}. In \bibinfo{booktitle}{\emph{Proceedingsof
  theSIGKDD Conference. Paris,France}}.
\newblock


\bibitem[Nabeel et~al\mbox{.}(2020)]%
        {graphbp}
\bibfield{author}{\bibinfo{person}{M. Nabeel}, \bibinfo{person}{I.~M. Khalil},
  \bibinfo{person}{B. Guan}, {and} \bibinfo{person}{T. Yu}.}
  \bibinfo{year}{2020}\natexlab{}.
\newblock \showarticletitle{Following Passive DNS Traces to Detect Stealthy
  Malicious Domains Via Graph Inference}.
\newblock \bibinfo{journal}{\emph{ACM Trans. Priv. Secur.}}
  \bibinfo{volume}{23}, \bibinfo{number}{4}, Article \bibinfo{articleno}{17}
  (\bibinfo{date}{July} \bibinfo{year}{2020}), \bibinfo{numpages}{36}~pages.
\newblock


\bibitem[Roberts et~al\mbox{.}(2019)]%
        {embeddingsquatting:2019}
\bibfield{author}{\bibinfo{person}{R. Roberts}, \bibinfo{person}{Y.
  Goldschlag}, \bibinfo{person}{R. Walter}, \bibinfo{person}{T. Chung},
  \bibinfo{person}{A. Mislove}, {and} \bibinfo{person}{D. Levin}.}
  \bibinfo{year}{2019}\natexlab{}.
\newblock \showarticletitle{You Are Who You Appear to Be: A Longitudinal Study
  of Domain Impersonation in TLS Certificates}. In
  \bibinfo{booktitle}{\emph{CCS}}. \bibinfo{pages}{2489--2504}.
\newblock


\bibitem[Schlichtkrull et~al\mbox{.}(2018)]%
        {rgcn}
\bibfield{author}{\bibinfo{person}{M. Schlichtkrull}, \bibinfo{person}{T.
  Kipf}, \bibinfo{person}{P. Bloem}, \bibinfo{person}{R. van¬†den Berg},
  \bibinfo{person}{I. Titov}, {and} \bibinfo{person}{M. Welling}.}
  \bibinfo{year}{2018}\natexlab{}.
\newblock \showarticletitle{Modeling Relational Data with Graph Convolutional
  Networks}. In \bibinfo{booktitle}{\emph{The Semantic Web}}.
  \bibinfo{pages}{593--607}.
\newblock


\bibitem[Shi et~al\mbox{.}(2017)]%
        {hin_survey:tkde:2017}
\bibfield{author}{\bibinfo{person}{Chuan Shi}, \bibinfo{person}{Yitong Li},
  \bibinfo{person}{Jiawei Zhang}, \bibinfo{person}{Yizhou Sun}, {and}
  \bibinfo{person}{Philip~S. Yu}.} \bibinfo{year}{2017}\natexlab{}.
\newblock \showarticletitle{A survey of heterogeneous information network
  analysis}.
\newblock \bibinfo{journal}{\emph{IEEE Transactions on Knowledge and Data
  Engineering}} \bibinfo{volume}{29}, \bibinfo{number}{1}
  (\bibinfo{year}{2017}), \bibinfo{pages}{17--37}.
\newblock
\urldef\tempurl%
\url{https://doi.org/10.1109/TKDE.2016.2598561}
\showDOI{\tempurl}


\bibitem[Silva et~al\mbox{.}(2021)]%
        {compromised}
\bibfield{author}{\bibinfo{person}{R. Silva}, \bibinfo{person}{M. Nabeel},
  \bibinfo{person}{C. Elvitigala}, \bibinfo{person}{I. Khalil},
  \bibinfo{person}{T. Yu}, {and} \bibinfo{person}{C. Keppitiyagama}.}
  \bibinfo{year}{2021}\natexlab{}.
\newblock \showarticletitle{Compromised or Attacker-Owned: A Large Scale
  Classification and Study of Hosting Domains of Malicious URLs}. In
  \bibinfo{booktitle}{\emph{{USENIX} Security}}. \bibinfo{pages}{3721--3738}.
\newblock


\bibitem[Sun et~al\mbox{.}(2019)]%
        {hindom}
\bibfield{author}{\bibinfo{person}{Xiaoqing Sun}, \bibinfo{person}{Mingkai
  Tong}, {and} \bibinfo{person}{Jiahai Yang}.} \bibinfo{year}{2019}\natexlab{}.
\newblock \showarticletitle{HinDom: A Robust Malicious Domain Detection System
  based on Heterogeneous Information Network with Transductive Classification}.
  In \bibinfo{booktitle}{\emph{RAID}}.
\newblock


\bibitem[Velickovic et~al\mbox{.}(2017)]%
        {gat}
\bibfield{author}{\bibinfo{person}{Petar Velickovic}, \bibinfo{person}{Guillem
  Cucurull}, \bibinfo{person}{Arantxa Casanova}, \bibinfo{person}{Adriana
  Romero}, \bibinfo{person}{Pietro Lio}, {and} \bibinfo{person}{Yoshua
  Bengio}.} \bibinfo{year}{2017}\natexlab{}.
\newblock \showarticletitle{Graph attention networks}.
\newblock \bibinfo{journal}{\emph{stat}}  \bibinfo{volume}{1050}
  (\bibinfo{year}{2017}), \bibinfo{pages}{20}.
\newblock


\bibitem[{VirusTotal, Subsidiary of Google}(2022)]%
        {virustotal}
\bibfield{author}{\bibinfo{person}{{VirusTotal, Subsidiary of Google}}.}
  \bibinfo{year}{2022}\natexlab{}.
\newblock \bibinfo{title}{{VirusTotal}}.
\newblock \bibinfo{howpublished}{\url{https://www.virustotal.com/}}.
\newblock


\bibitem[Weimer(2005)]%
        {Weimer:2005:PDNS}
\bibfield{author}{\bibinfo{person}{Florian Weimer}.}
  \bibinfo{year}{2005}\natexlab{}.
\newblock \showarticletitle{{Passive DNS Replication}}. In
  \bibinfo{booktitle}{\emph{FIRST}}. \bibinfo{pages}{98}.
\newblock


\end{thebibliography}


\end{document}